# Building Computational Models to Predict One-Year Mortality in ICU Patients with Acute Myocardial Infarction and Post Myocardial Infarction Syndrome

Laura A. Barrett, BA[1], Seyedeh Neelufar Payrovnaziri, MS[1],
Jiang Bian, PhD[2], Zhe He, PhD[1]
[1]School of Information, Florida State University, Tallahassee, Florida, USA;
[2]Department of Health Outcomes and Biomedical Informatics, University of Florida, Gainesville, Florida, USA

**Abstract**

*Heart disease remains the leading cause of death in the United States. Compared with risk assessment guidelines that require manual calculation of scores, machine learning-based prediction for disease outcomes such as mortality can be utilized to save time and improve prediction accuracy. This study built and evaluated various machine learning models to predict one-year mortality in patients diagnosed with acute myocardial infarction or post myocardial infarction syndrome in the MIMIC-III database. The results of the best performing shallow prediction models were compared to a deep feedforward neural network (Deep FNN) with back propagation. We included a cohort of 5436 admissions. Six datasets were developed and compared. The models applying Logistic Model Trees (LMT) and Simple Logistic algorithms to the combined dataset resulted in the highest prediction accuracy at 85.12% and the highest AUC at .901. In addition, other factors were observed to have an impact on outcomes as well.*

**Introduction**

Heart disease is the leading cause of death in the United States. About half of Americans have at least one of the three most common risk factors for heart disease: high blood pressure, high low-density lipoprotein (LDL) cholesterol, and smoking. Other factors that contribute to cardiovascular health are physical activity, diet, weight, and glucose control. By 2035, it is estimated that almost half of adults in the United States will have some form of heart disease, with the cost exceeding $1.1 trillion.[1] Around 720,000 Americans will have their first hospitalization due to acute myocardial infarction (AMI) or coronary heart disease, and around 335,000 will have a recurrent event, in 2018.[1] It was suggested that one in every 7.4 Americans will die of an AMI.[1] Moreover, 170,000 AMI incidences, out of approximately 805,000 every year, are silent or without the classic symptoms such as chest pain, shortness of breath, and digestive issues.[1] Most treatment plans involve lifestyle modification and medications that require a high level of patient participation, compliance, and agreement with the treatment plan. Non-compliance has been studied and the results remain consistent that compliance is key to reduced cost-benefit, lower readmission rates, and reduced mortality.[2–4] An accurate prediction model for AMI mortality can inform the design of treatment plans and promise to improve patient compliance with the plans, thereby reducing costs, readmission, and mortality rates. In addition, it can set realistic expectations for medical staff, patients, and families at a time when emotion and judgment conflict. An acute medical illness that requires hospitalization often results in a high mortality rate during the year after discharge, especially for elderly patients. This is due to progressive physical decline.[5] Considering severely ill patients, who are usually prone to unnecessary interventions, communication about end-of-life planning (EOLp) becomes critical to the hospitals. A timely and reasonable discussion on EOLp is possible when there is an accurate estimation of patient's risk of death.[6]

Currently there are a variety of score-based risk assessment guidelines and systems associated with mortality prediction for outcomes of AMI and other heart diseases. CADILLAC (Controlled Abciximab and Device Investigation to Lower Late Angioplasty Complications) risk score is used to predict mortality for patients that suffered from an ST-segment-elevation myocardial infarction (STEMI) with percutaneous coronary intervention (PCI) as the primary treatment.[7] CRUSADE (Can Rapid risk stratification of Unstable angina patients Suppress ADverse outcomes with Early implementation of the ACC/AHA guidelines) is a risk assessment scoring system that measures bleeding risk in patients with non-ST-segment-elevation myocardial infarction (NSTEMI).[8] Global Registry of Acute Coronary Events (GRACE) is used to predict in-hospital and six-month mortality risk following an AMI.[7] Korea Acute Myocardial Infarction Registry is a prediction model for long term outcomes related to NSTEMI.[9] The Ontario mortality prediction scores can be used to predict 30-day and one-year mortality rates following an AMI.[10] The Primary Angioplasty in Myocardial Infarction (PAMI) mortality risk score, which is similar to CADILLAC but uses different variables, is specifically used when primary PCI was the treatment of choice for patients who suffered from a STEMI.[11] The Soroka Acute Myocardial Infarction (SAMI) score predicts one- and five- year mortality following an AMI.[12] The Seattle Post Myocardial Infarction Model (SPIM) can predict survival following an AMI with left ventricular dysfunction.[13] Thrombolytic In Myocardial Infarction (TIMI) risk score is to predict in-hospital and one-year mortality in STEMI patients.[7] The Zwolle risk score is used for early discharge of low-risk patients following primary PCI for patients that suffered from a STEMI.[14]

There is not one go-to risk scale for all AMI patients, since it depends on the specific situation of the patient. In addition, different kinds of information are required for different risk scores. Though calculators do exist for some of these scales, the required information may not always be readily available. For example, Killip class is not readily available in the electronic health records (EHR), but is needed for calculating a majority of aforementioned scores. Killip class, is a major factor considered in many scores, where classes are based on physical examination of the patient to gauge the level of heart failure.[15] Other factors used in these different scoring systems include demographic information such as age, gender, weight, ethnics; chart values such as heart rate and systolic blood pressure; lab values such as hematocrit, creatinine clearance, serum and creatinine; test results such as echocardiography results, TIMI flow rate, and left ventricular ejection fraction; type, timing, and location of AMI; treatment decision; cardiac health information including presence of triple vessel disease, cerebrovascular disease, pulmonary edema, cardiac dysrhythmias, angina, hypertension; and information regarding comorbidities and overall health such as diabetes status, anemia, renal health, cancer status, smoking status, and obesity.[7-14]

With the wide adoption of EHR systems in the US, the rich EHR data have been recognized as a promising resource for comparative effectiveness research, outcomes research, epidemiology, drug surveillance, and public health research.[16] On the other hand, machine learning has become a popular approach for building models using EHR data to predict mortality rates, as well as a variety of other disease outcomes such as progression to Type 2 diabetes,[17] intensive care readmission,[18] and development of Alzheimer's disease.[19] There are other examples of predicting outcomes for radiation therapy[20] and acute ischemic stroke.[21] Lately, multiple studies have used machine learning and EHR data to predict mortality for patients diagnosed with diabetes,[22] breast cancer,[23] burn injury,[24] and sepsis or septic shock.[25-26] The information availability and ease of access to information within EHR has allowed for this newer technology to move to the forefront. When previous predictive models were built, they were done so when paper files were still the standard. Since EHRs have now become the *de facto* data source for health-related research, it follows that utilizing machine learning to build prediction models would also become the new standard.

This is especially true, in cardiology, as practitioners become required to utilize large amounts of data from different sources and practice more personalized medicine instead of a one-size fits all approach.[27,28] In addition, the complexity of these patients can create an environment well suited for machine learning techniques. Recently, a few studies have used machine learning to predict cardiovascular issues. Goldstein, Navar and Carter[29] worked specifically in the pros and cons of using machine learning versus regression models in cardiovascular risk prediction. They studied 1,944 patients with AMI admitted through the emergency room and utilized lab tests, demographics, and comorbidity information. Wallert et al.[30] compared four different machine learning algorithms on 51,943 patients to predict two-year survival following an AMI. They evaluated boosted C5.0 trees/rule sets, Random Forests, Support Vector Machines (SVM), and Logistic Regression. They utilized 39 different predictors, such as established risk indicators, survival factors, demographics, symptoms, echocardiography results, lab values, reperfusion, and pharmacology. Mansoor et al.[31] studied in-hospital mortality in females with STEMI, using Logistic Regression and Random Forest algorithms. Their study consisted of 12,047 patients and utilized a different number of variables, including demographic information, comorbidities, treatment, health statics, zip code, primary payer, admission information, and hospital characteristics. Miao et al.[32] predicted one-year mortality in cardiac arrhythmias. They studied 10,488 patients from MIMIC-II using 40 variables and Random Survival Forest classification. Their variables included demographics, clinical variables, lab values, and pharmacology. Deep learning and artificial neural networks, have recently been widely used in prediction tasks relating to healthcare due to their superior prediction accuracy.[33,34] Auto-encoders have successfully been utilized to predict diagnosis. Recurrent neural networks (RNN) have been applied to handle temporal sequence of events.[33]

The current study built and evaluated various machine learning models to predict one-year mortality in patients that were diagnosed with an AMI or post myocardial syndrome (PMS). One-year mortality was selected for this study as a starting point because it would allow for comparison to other studies and consider that there are patients that had multiple AMI admissions within a two-year period. In our study, the results of the best performing machine learning model are compared to a deep feedforward neural network (Deep FNN) with back propagation. There are some overlapping characteristics compared to previous cardiac associated prediction models and some differences. For example, this study does not differentiate between NSTEMI and STEMI as many previous models do. In addition, this study does not use any interpreted data such as echocardiography, location of AMI, or Killip class. The goal is to build prediction models using different machine learning algorithms and datasets to determine which classification model(s) performs the best. As shown by Johnson et al. there are pros and cons in each of the algorithms.[27] Based on the results from many recent studies, it is proved that machine learning models are more accurate in mortality prediction tasks compared to score-based systems. Contrary to score-based guidelines, which require physicians to manually calculate scores, machine learning-based models for predicting disease outcomes are based on complex algorithms with powerful analysis capability on large datasets.[32]

**Methods**

We obtained patient data from the Medical Information Mart for Intensive Care III (MIMIC-III) database.[36] MIMIC-III is a freely available database developed by the MIT Lab for Computational Physiology. It integrates deidentified, comprehensive clinical data of patients admitted to the Beth Israel Deaconess Medical Center in Boston, Massachusetts. It is released as a relational database comprised of 26 data tables of which 7 were used in this study. This database covers over 58,000 hospital admissions of approximately 40,000 patients spanning an eleven-year period.[36] Waikato Environment for Knowledge Analysis (WEKA) is a freely available Java based software developed at the University of Waikato, New Zealand. It provides user-friendly and easy-to-use implementations of many machine learning algorithms.[37] WEKA was used in this study to explore the performance of various machine learning algorithms on the obtained datasets. In WEKA, we used the default configuration of each algorithms with 10-fold-cross-validation for model validation. TensorFlow is an open-source machine learning system released by Google Brain.[38] It is used in this study to implement a Deep FNN, utilizing packages from SciPy,[39] including Scikit-learn,[40] NumPy[41] and Pandas[42] in Python,[43] on a system with 2.2 GHz Intel Core i7 CPU. The implemented Deep FNN in this study has 2 hidden layers and 1 prediction layer for the classification task, using softmax cross entropy with logits for cost estimation, and gradient descent algorithm to optimize the cost. The size of each hidden layer was set as twice the input size. Rectified linear unit (ReLU) was used as the linear activation function. 10-fold-cross-validation was considered for model evaluation since data is not very large.[44]

*Data and Variable Selection*

Data selection was initially based on features used in other similar studies and previous risk scales, and then further refined and limited by the data available in MIMIC-III. Interpreted data such as echocardiography and Killip class, certain specific comorbidity diagnoses, and pharmacology were not readily available in MIMIC-III and excluded because of that. Features related to admission, demographic, treatment, and comorbidity, were considered in this study. Regarding lab values, those evaluating both long- and short-term overall health, were taken into consideration. In addition, kidney and/or liver function have proven to be predictors of mortality,[7–9,13,22,25,26] so values related to these were included. Lastly, since preferred cardiac markers changed over the eleven-year period, a variety of them were included as well. Any data value with multiple entries such as lab values were averaged and the average value was used. In this study, we created different real-world datasets (i.e., admission information, demographics, treatment information, diagnostic information, and lab & chart values) from MIMIC-III, and investigated which set of features would contribute to the most reliable prediction models. In addition, a combination of all datasets was created to see if it improves model performance. Details regarding the variables employed in this study are given in Table 1.

Initially, there were 5037 subjects that met the criteria to be included in this study. These were patients who had an ICD-9 code of 410.0-411.0 (used for claims with a date of service on or before September 30, 2015) in MIMIC-III. Upon further review, it was determined that these subjects accounted for a total of 7590 admissions over an eleven-year period, out of which only 5436 admissions had a diagnosis of AMI or PMS, because there were patients that had multiple admissions for AMI or PMS while others had multiple admissions with only one being for AMI or PMS. It was decided to look at the data based on admission, rather than the individuals, since there were cases in which an individual survived a year in one admission, but did not survive a year in another admission. We treated each admission as a separate instance because each admission presented set of circumstances. Though certain demographic features would remain the same if the same patient was readmitted, all of the other factors could be different.

Table 1. Variables included in each dataset

| Dataset | Variables |
|---|---|
| Admission Information | Total days of admission, month of admission, discharge location, whether or not initial ER diagnosis was AMI or rule out AMI |

| Demographics | Age at admission, gender, religion, ethnicity, marital status |
|---|---|
| Treatment Information | Cardiac catheterization, cardiac defibrillator and heart assist anomaly, cardiac defibrillator implant with cardiac catheterization, cardiac defibrillator implant without cardiac catheterization, cardiac valve and other major cardiothoracic procedures with cardiac catheterization, cardiac valve and other major cardiothoracic procedures without cardiac catheterization, cardiac valve procedures with cardiac catheterization, cardiac valve procedures without cardiac catheterization, coronary bypass with cardiac catheterization, coronary bypass with cardiac catheterization or percutaneous cardiac procedure, coronary bypass with PTCA, coronary bypass without cardiac catheterization, coronary bypass without cardiac catheterization or percutaneous cardiac procedure, other cardiac pacemaker implantation, other major cardiovascular procedures, other permanent cardiac pacemaker implant or PTCA with coronary artery stent implant, percutaneous cardiac procedure with drug-eluting stent, percutaneous cardiac procedure with non-drug-eluting stent, percutaneous cardiac procedure without coronary artery stent, percutaneous cardiovascular procedure, and permanent cardiac pacemaker implant |
| Diagnostic Information | Cancer, endocrinology, gastroenterology, genitourinary, hematological disorder, infection, liver or kidney issues, neurological disorder, orthopaedic disorder, other cardiovascular disease, other comorbidities (DRG codes), respiratory disorder, and toxicity issues |
| Lab & Chart Values | Average lab values included a lipid profile (cholesterol ratio, LDL cholesterol, HDL cholesterol, total cholesterol, and triglycerides), liver and kidney function tests (alanine transaminase, aspartate transaminase, alkaline phosphatase, albumin, bilirubin, blood urea nitrogen, creatinine, gamma-glutamyltransferase, L-lactate dehydrogenase, and total protein), cardiac function tests (N-terminal prohormone of B-type natriuretic peptide, C-reactive protein, creatine kinase, creatine phosphokinase-MB, cortisol, homocysteine, troponin I and troponin T), and electrolytes (bicarbonate, calcium, chloride, potassium, and sodium), glucose, hematocrit, hemoglobin, and white blood count; Average chart values include diastolic and systolic blood pressure, heart rate, and respiratory rate |
| Combined | All of the Above |

*Data Preprocessing*

From the initial list of hospital admissions, there were admissions that had multiple diagnostic codes for AMI or AMI and PMS. Duplicates were removed to ensure that there was one admission per actual instance. Duplicate entries also existed in the treatment data and the comorbidities. In both cases, it was either due to multiple codes for the same treatment (due to multiple entries in different parts of the EHR) or due to multiple treatments or comorbidities for the same admission. In the latter case, both items were counted. For the demographics, one data item that needed to be corrected was the age and death age of anyone over the age of 89. These fields were changed in MIMIC-III to mask the age per HIPAA rules. The ages were changed back by subtracting 211 years from the values presented. The lab and chart values were reviewed to remove or correct entry errors. There were a few instances in the blood pressures where the numbers had been entered in reverse order. It was also realized that a '0' value in the lab values occur when there is a 'see comment' entered at any time with that lab value, thus all 0 values were removed from the lab values. Any lab or chart values that seemed biologically invalid were also removed. Outliers in lab values were removed. The first and third quartile were calculated and any values that fell outside of the first quartile minus 1.5 times the interquartile range or the third quartile plus 1.5 times the interquartile range were removed. Classifications were run with both the outliers in place and without the outliers. Results were better with the outliers. This was potentially due to more abnormal lab results.

*Modeling and Evaluation*

The outcome of the prediction models is the mortality that occurred within one year after admission. Admissions in which the patient died within a year were considered as positive instances and those in which the patient did not die within a year were considered as negative instances. The classification algorithms that were run for each dataset include AdaBoost, Attribute Selected Classifier, Bayes Net, Classification Via Regression, Decision Stump, Decision Table, Iterative Classifier Optimizer, Hoeffding Tree, J48, JRip, Logistic, Logistic Model Trees (LMT), LogitBoost, NaiveBayes, OneR, PART, Random Forest, Random SubSpace, Random Tree, Randomizable Filtered Classifier, REP Tree, Stochastic Gradient Descent (SGD), Simple Logistic, Sequential Minimum Optimization (SMO), and Voted Perceptron. As some algorithms performed poorly, not all results are reported here. The classification algorithms were compared to determine which models yielded the best results based on overall accuracy, precision, recall, and area under the curve (AUC) of the receiver operator curve (SROC). Precision measures positive versus negative results, while recall measures the percent of positive instances identified out of the total amount of positive instances in the dataset. F-measure is the harmonic mean of precision and recall. Machine learning algorithms were first applied on each dataset individually, and then on the combined dataset with predicted outcome as one-year mortality. Deep FNN was applied on the combined dataset and compared with the best performing shallow machine learning algorithm.

**Results**

Of the 5436 admissions included in this study, 1629 (30%) in which the patient died within one year of admission are positive instances and the remaining 3807 (70%) in which the patient survived past the one-year mark are classified as negative instances. Summary statistics for these instances can be found in Table 2.

Table 2. Summary statistics of the cohort of patients

| Characteristics | | Number of Instances | Number and Percentage of Positive Instances | | Number and Percentage of Negative Instances | | P-value |
|---|---|---|---|---|---|---|---|
| Gender | Male | 3308 | 877 | 26.5% | 2431 | 73.5% | <.00001 |
| | Female | 2128 | 752 | 35.3% | 1376 | 64.7% | <.00001 |
| Age when first admitted | Under 30 | 13 | 4 | 30.8% | 9 | 69.2% | .949588 |
| | 30 to 49.9 | 420 | 58 | 13.8% | 362 | 86.2% | <.00001 |
| | 50 to 59.9 | 784 | 92 | 11.7% | 692 | 88.3% | <.00001 |
| | 60 to 69.9 | 1209 | 273 | 22.6% | 936 | 77.4% | <.00001 |
| | 70 to 79.9 | 1428 | 465 | 32.6% | 963 | 67.4% | .012632 |
| | 80 to 90 | 1524 | 708 | 46.5% | 816 | 53.5% | <.00001 |
| | Over 90 | 58 | 29 | 50.0% | 29 | 50.0% | .000813 |
| Ethnicity | Asian | 87 | 23 | 26.4% | 64 | 73.6% | .468718 |
| | Black | 291 | 103 | 35.4% | 188 | 64.6% | .037736 |
| | Hispanic/Latino | 97 | 19 | 19.6% | 78 | 80.4% | .024348 |
| | Other | 108 | 25 | 23.1% | 83 | 76.9% | .118186 |
| | White | 3774 | 1144 | 30.3% | 2630 | 69.7% | .401699 |
| Initial ER Diagnosis MI? | Yes | 2061 | 470 | 22.8% | 1591 | 77.2% | <.00001 |
| | No | 3375 | 1159 | 34.3% | 2216 | 65.7% | <.00001 |
| With Comorbidities | Total | 1614 | 798 | 49.4% | 816 | 50.6% | <.00001 |

Figure 1 shows the ROC for the datasets based on the classification that produced the highest accuracy for each dataset. The best performing models, in terms of AUC, were generated by applying Logistic Model Trees (LMT) and Simple Logistic algorithms on the combined dataset. Applying J48 and Logistic Regression algorithms, on the admission dataset, and the lab & chart values dataset, respectively, resulted in models with .80 AUC. Models that resulted from applying Random Forest, Attribute Select Classifier, and Decision Table algorithms on the treatment, demographics, and comorbidities datasets, respectively, did not perform well, all scoring an AUC under .70.

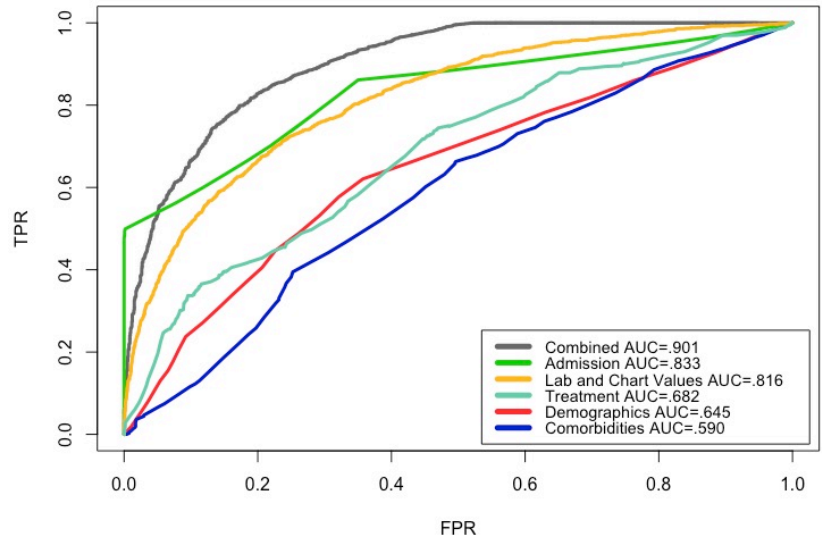

**Figure 1.** ROC of shallow models based on the highest accuracy with each dataset.

Table 3 shows the results based on highest accuracy for each dataset, along with the shallow classification algorithm used. The model based on the combined dataset showed the highest accuracy of 85.12% applying LMT and Simple Logistic algorithms. The model developed from J48 algorithm on the admission dataset was closely behind at 84.88% accuracy. These two models were not drastically different in other measurements as well. The model developed from J48 algorithm on the admission dataset, resulted in the highest precision, meaning it returned the most positive versus negative instances. The recall rates were low across the board meaning that none of these models were especially great at returning a high percentage of positive instances. In the case of the comorbidities dataset, modeled using Decision Table algorithm, the sensitivity is slightly higher than the specificity meaning that this dataset and model were slightly better at detecting the positive instances than the negative instances.

**Table 3.** Summary of WEKA results based on the highest accuracy

| Dataset | Best Performing Classification Algorithm | Highest Percent Correctly Classified | AUC | Precision | Recall | F-Measure |
|---|---|---|---|---|---|---|
| Combined | LMT and Simple Logistic | **85.12%** | **0.901** | 0.867 | 0.594 | 0.705 |
| Admission | J48 | 84.88% | 0.833 | **0.993** | 0.499 | 0.664 |
| Treatment | Random Forest | 83.73% | 0.682 | 0.789 | 0.028 | 0.055 |
| Lab & Chart Values | Logistic | 78.46% | 0.816 | 0.708 | 0.478 | 0.571 |
| Demographics | Attribute Selected Classifier | 70.70% | 0.645 | 0.524 | 0.238 | 0.327 |
| Comorbidities | Decision Table | 58.18% | 0.590 | 0.566 | 0.660 | 0.610 |

Table 4 shows the results of applying various shallow machine learning algorithms in WEKA and the deep learning algorithm implemented in TensorFlow, on the combined dataset. For the most part, the models based on machine learning algorithms performed similarly well with a few exceptions. LMT and Simple Logistic showed the highest accuracy of 85.12%. They also showed the highest AUC at .901. Decision Stump had the highest precision at 1 meaning 100% of instances classified as positive were indeed positive. None of the machine learning based models showed recall rates that would be considered stellar, the highest rate being .744 produced by Bayes Net. This shows that none of these models were great at detecting a high percentage of positive instances in the dataset. This is where we examined Deep FNN. The model using Deep FNN, drastically outperformed all machine learning algorithms in terms of recall at .820, while maintaining precision at 0.831 and F-measure at .813. This means the Deep FNN model outperformed the shallow machine learning algorithms in identifying a higher percentage of positive instances. Since its AUC is .751, this model does not have a great sensitivity or specificity and its ROC curve would fall somewhere between the ones for lab & chart values and treatment in Figure 1.

Table 4. Comparison of the performance of different models using the combined dataset

| Classification Algorithms | Percent Correctly Classified | AUC | Precision | Recall | F-Measure |
|---|---|---|---|---|---|
| AdaBoost | 84.28% | 0.849 | 0.981 | 0.484 | 0.649 |
| Bayes Net | 81.55% | 0.889 | 0.674 | **0.744** | 0.707 |
| Decision Stump | 84.31% | 0.730 | **1** | 0.476 | 0.645 |
| Decision Table | 84.29% | 0.865 | 0.923 | 0.519 | 0.665 |
| J48[a] | 84.64% | 0.843 | 0.97 | 0.503 | 0.662 |
| JRip | 84.27% | 0.737 | 0.985 | 0.483 | 0.648 |
| LMT | **85.12%** | **0.901** | 0.867 | 0.594 | 0.705 |
| Logistic | 85.06% | 0.899 | 0.835 | 0.626 | **0.715** |
| Naïve Bayes | 82.63% | 0.862 | 0.768 | 0.603 | 0.675 |
| OneR | 84.88% | 0.749 | 0.993 | 0.499 | 0.664 |
| PART | 84.58% | 0.869 | 0.973 | 0.500 | 0.660 |
| Random Forest | 84.60% | 0.893 | 0.949 | 0.514 | 0.667 |
| Random Tree | 76.77% | 0.776 | 0.628 | 0.552 | 0.588 |
| REP Tree | 83.61% | 0.845 | 0.884 | 0.522 | 0.656 |
| SGD | 84.90% | 0.765 | 0.904 | 0.555 | 0.688 |
| Simple Logistic | **85.12%** | **0.901** | 0.867 | 0.594 | 0.705 |
| SMO[b] | 84.84% | 0.751 | 0.972 | 0.509 | 0.668 |
| Voted Perceptron | 70.70% | 0.519 | 0.773 | 0.031 | 0.06 |
| Deep FNN | 82.02% | 0.751 | 0.831 | **0.820** | **0.813** |

[a]: The implementation of C4.5 decision tree in WEKA
[b]: WEKA's implementation of John Platt's sequential minimal optimization algorithm for training a support vector classifier.

Table 5 shows a comparison of the results of this study to similar cardiovascular studies. As shown, this study resulted in the highest AUC. It is interesting that no study utilized the same machine learning algorithm to obtain the highest AUC in their best model. This reflects a common understanding in data science that there is no universally applicable algorithm that outperforms all the other algorithms all the time.

Table 5. Comparison of AUC to other Cardiovascular Studies

| Study | Study Topic | Classification Algorithm | Highest AUC | # of Variables |
|---|---|---|---|---|
| Barrett et al. (This paper) | 1-year mortality AMI | Simple Logistic/LMT | 0.901 | 79 |
| Goldstein et al.[29] | Cardiovascular Risk | Boosting Tree Based | 0.763 | 43 |
| Mansoor et al[31]. | In-hospital mortality STEMI | Logistic Regression | 0.840 | 11 (with backward elimination) |
| Miao et al.[32] | 1-year mortality arrythmia | Random Survival Forest | 0.810 | 40 |
| Wallert et al.[30] | 2-year survival AMI | Support Vector Machine | 0.845 | 39 |

**Discussion**

There are many factors that can affect mortality rates following a myocardial infarction. Figuring out a way to utilize the information regarding these factors will assist in accurately predicting possible outcomes. As seen from this dataset, there is not one specific factor that provides the needed predictability information, while being able to include all relevant criteria leads to improved predictions. The various factors including comorbidities, treatment options, patient history, short-term and long-term overall health, and demographics can all impact outcomes in cardiac health, which makes it important to consider all of them when predicting cardiac outcomes.

One interesting finding in this study, was the importance of initial diagnosis, as a "myocardial infarction" or "rule out myocardial infarction". Of the 5436 admissions, 3375 of these were not tagged as either of these as the primary reason for admission. Of these, the percentage in the positive instances was higher than the percentage in the positive instances (34.3% versus 30%). Chi-square statistics was applied to compare the expected outcome with the observed outcome. The p-value was < .00001 and the result was significant at p < .05. This indicates the importance of recognizing myocardial infarction initially so that appropriate treatment can be used as soon as possible. Another possible explanation for this is when patient was hospitalized for a comorbidity, a significant number of AMI occurred. As there is no way to time stamp when the AMI occurred for patients who had AMI before admission, it is not possible to differentiate between those that occurred prior to admission and after admission. If that is the case, it indicates that an AMI that occurs during an admission may lead to a higher mortality rate.

Two elements of the demographic dataset also appeared to be statistically significant. First off was gender. The percentage of female was higher than the percentage of male in the positive class. Chi-square statistics, considering. p-value < .00001 and significance at p < .05, indicate that gender does play a role in one-year mortality following a myocardial infarction. This may be attributed to differing symptoms in male and female patients. Because of the difference of symptoms in males and females, perceptional differences about these symptoms can play a role. This may cause female to seek treatment less proactively than male due to downplaying the symptoms or attributing the cause of the symptoms to something else. Another factor could be age, the average age of male in the positive class was 75.08 versus the average age of female in the same class at 77.65. In the negative class the gap is greater with males at an average of 66.02 and females averaging 72.09. The second element was ethnicity. Those that identified as White and Black had a higher percentage in the positive class than the overall cohort did. Those that identified as Asian and Hispanic/Latino had a lower percentage in the positive class than the overall cohort did.

Admissions in the comorbidity dataset had a higher percentage in the positive class than the overall cohort did. Statistical analysis indicates that if patients have a comorbidity on the same admission, they are more prone to die within a year of a myocardial infarction than someone that does not have any treatment for a comorbidity on the same admission.

*Limitations and Future Opportunities*

A potential limitation of this study was the imbalanced dataset of positive and negative cases (30% vs. 70%). Future studies could be conducted in which there is closer to a 50/50 balance using some random sampling strategies. Another limitation was the availability of some lab values in MIMIC-III. Some patients did not have all lab values collected. In addition, some of the desired lab and chart values could not be easily obtained. Potentially, a different database would have different access to lab values and chart values allowing for more robust records to be used. Another limitation to this study is that none of the machine learning models and Deep FNN model were compared against the standard predictive scores to determine which were more effective and efficient. This would be an interesting further study if the variables needed for calculating the risk scores (e.g., Ontario, SAMI) are available in EHR. One last limitation to this study is the lack of information related to cause of death. We are unable to tell if cause of death was related to AMI, cardiac health, a comorbidity, or a completely unconnected occurrence. This additional information may provide helpful insight into improving mortality rates.

Other opportunities for additional studies include those related to mortality rates as they relate to correct identification of AMI in the ER versus rates of those that have an AMI while already admitted for an unrelated comorbidity. Based on the data system in MIMIC-III, there is no specific timeline associated with when the AMI occurred so it is not possible to note if the diagnosis was associated with the initial admission or occurred during an unrelated ICU stay. In addition, studying the role of gender feature in mortality related to AMI would help. This study noted that the mortality rate in female was higher than male, thus discovering the reasons would be helpful to lowering those rates. Further studies could also use differing timelines related to mortality rates, for example 30 days, six months, and two-years as done in other comparable studies. The deep learning architecture that was used in this study was a fully connected feedforward neural network. The non-linear activation function considered for hidden layers is usually logistic sigmoid, tanh, or ReLU. We chose ReLU which yielded the best performance results. During the training, the cross-entropy loss was optimized between the target output and prediction output. Another future study could investigate other deep learning algorithms' performances on mortality prediction, like convolutional neural network (CNN), recurrent neural network (RNN), or long short-term memory (LSTM), incorporating more dense features (e.g., longitudinal data) and records, including clinical notes and temporal information related to patients.

## Conclusions

This study examined multiple machine learning algorithms on multiple datasets obtained from MIMIC-III, to predict one-year mortality rate in patients diagnosed with AMI or PMS. In addition, the results of the best performing machine learning model on the Combined dataset, was compared to a Deep FNN model. Highest prediction accuracy and AUC was achieved by LMT and Simple Logistic algorithms, while Deep FNN had less accuracy and precision but higher recall. Correct diagnosis and treatment of AMI can have an effect on mortality as this study has shown. This emphasizes the importance of a few things. One, there needs to be more training to the public on reducing the risk factors associated with heart disease and AMI. Two, in order to lower the chances of a comorbidity, there needs to be more incentives to the public to maintain overall health when risk factors of heart disease exist. Three, there also needs to be more training to the public on recognizing the signs and symptoms of AMI, especially in women. Four, the improved predictability obtained by using machine learning can help at-risk patients strive for compliance to treatment plans to improve mortality risk.